# Influence of Early through Late Fusion on Pancreas Segmentation from Imperfectly Registered Multimodal MRI

**Lucas W. Remedios[a]\***, **Han Liu[a]**, **Samuel W. Remedios[c,d]**, **Lianrui Zuo[b]**, **Adam M. Saunders[b]**, **Shunxing Bao[b]**, **Yuankai Huo[a,b]**, **Alvin C. Powers[j,k,l]**, **John Virostko[f,g,h,i]**, **Bennett A. Landman[a,b,e]**

[a]Vanderbilt University, Department of Computer Science, Nashville, USA
[b]Vanderbilt University, Department of Electrical and Computer Engineering, Nashville, USA
[c]Johns Hopkins University, Department of Computer Science, Baltimore, USA
[d]National Institutes of Health, Department of Radiology and Imaging Sciences, Bethesda, USA
[e]Vanderbilt University, Department of Biomedical Engineering, Nashville, USA
[f]University of Texas at Austin, Department of Diagnostic Medicine, Dell Medical School, Austin, USA
[g]University of Texas at Austin, Livestrong Cancer Institutes, Dell Medical School, Austin, USA
[h]University of Texas at Austin, Austin, Department of Oncology, Dell Medical School, USA
[i]University of Texas at Austin, Oden Institute for Computational Engineering and Sciences, Austin, USA
[j]Vanderbilt University Medical Center, Department of Medicine, Division of Diabetes, Endocrinology, and Metabolism, Nashville, USA
[k]VA Tennessee Valley Healthcare System, Nashville, USA
[l]Vanderbilt University, Department of Molecular Physiology and Biophysics, Nashville, USA

**Abstract**

**Purpose:** Combining different types of medical imaging data (multimodal fusion) promises better segmentation of anatomical structures, such as the pancreas. Strategic implementation of multimodal fusion could improve our ability to study diseases like diabetes. However, where to perform fusion in deep learning models is still an open question. It is unclear if there is a single best location to combine (fuse) information when analyzing pairs of imperfectly aligned images, or if the optimal fusion location depends on the specific model being used. Two main challenges when using multiple imaging modalities to study the pancreas are 1) the pancreas and surrounding abdominal anatomy have a deformable (changeable) structure, making it difficult to consistently align the images and 2) breathing by the individual during image collection further complicates the alignment between multimodal images. Even after using state-of-the-art deformable image registration techniques, specifically designed to align abdominal images, multimodal images of the abdomen are often not perfectly aligned. This study examines how the choice of different fusion points, ranging from early in the image processing pipeline to later stages, impacts the segmentation of the pancreas on imperfectly registered multimodal magnetic resonance (MR) images.

**Approach:** Our dataset consists of 353 pairs of T2-weighted (T2w) and T1-weighted (T1w) abdominal MR images from 163 subjects with accompanying pancreas segmentation labels drawn mainly based on the T2w images. Because the T2w images were acquired on two breath holds (interleaved) and the T1w images on one breath hold, there were 3 different breath holds impacting the alignment of each pair of images. We used state-of-the-art deformable abdominal image registration (deeds) to align the image pairs. Then we trained a collection of basic UNets with different fusion points, spanning from early to late (early layers or late layers in the model), to assess how early through late fusion influenced segmentation performance on imperfectly aligned images. To investigate whether performance differences on key fusion points generalized to other architectures, we expanded our experiments to nnUNet.

**Results:** The single-modality T2w baseline using a basic UNet model had a Dice score of 0.73, while the same baseline on the nnUNet model achieved 0.80. For the basic UNet, the best fusion approach occurred in the middle of the encoder (early/mid fusion), which led to a statistically significant improvement of 0.0125 on Dice score compared to the baseline. For the nnUNet, the best fusion approach was naïve image concatenation before the model (early fusion), which resulted in a statistically significant Dice score increase of 0.0021 compared to baseline.

**Conclusions:** Fusion in specific blocks can improve performance, but the best blocks for fusion are model specific, and the gains are small. In imperfectly registered datasets, fusion is a nuanced problem, with the art of design remaining vital for uncovering potential insights. Future innovation is needed to better address fusion in cases of imperfect alignment of abdominal image pairs. Code associated with this project is available here: https://github.com/MASILab/influence_of_fusion_on_pancreas_segmentation.

**Keywords**: multimodal, fusion, pancreas segmentation, registration, MRI, UNet, nnUNet

*Lucas W. Remedios**, E-mail: lucas.w.remedios@vanderbilt.edu

# 1 Introduction

Multimodal fusion is the combination of information from different modalities with the aim of improving understanding of a scene, such as combining information from natural images and infrared images[1]. In medical imaging, multimodal fusion is combining information from multiple modalities, including imaging and tabular data, to provide a more complete assessment of subject condition based on complementary information[23]. In pure multimodal medical image fusion, combining imaging modalities that extract different anatomical information of the same subject is a rational approach, like leveraging both computed tomography images (for bone), and magnetic resonance (MR) images (for soft tissue characterization)[4]. In some medical imaging problems, it can be useful to combine several types of MR images (different modalities), as has been shown in the Brain Tumor Segmentation Challenge (BraTS), where T1w, contrast enhanced T1w, T2w, and fluid-attenuated recovery T2w (FLAIR) MR images were fused to provide a more complete analysis of brain tumors[5].

In this work, the disease of interest is diabetes, which greatly affects patients' lives and can lead to complications with vision, kidneys, and even lower limb amputation[6,7]. Diabetes is a global health crisis, with an estimated global prevalence of 10.5% in 2021[8]. The pancreas has two major compartments, endocrine and exocrine, with the insulin-secreting pancreatic islets of the endocrine compartment comprising about 2% of the pancreas volume. Pancreatic islet dysfunction plays a major role in most forms of diabetes.[9]. The total pancreatic volume is reduced in type 1 diabetes[10–13], indicating that the exocrine compartment is also altered. For these reasons, segmentation of the pancreatic images is relevant to understanding the changes in the pancreas in diabetes. Additionally, pancreas segmentation may be useful in the research of other diseases, like pancreatic cancer[14].

Unfortunately, the pancreas is notoriously difficult to segment due to its variable shape and lack of distinct boundaries in medical images[15,16]. It is sensible to leverage multiple imaging modalities that provide complementary information about structures around the pancreas to improve pancreas segmentation. For example, leveraging T2w MR images (delineating the stomach and filled intestines) and T1w MR images (delineating the liver and empty intestines) is a reasonable approach for better determining the pancreas boundary.

Generally, the first step in the fusion of multimodal medical images is registration so that the anatomical structures are aligned across modalities[17]. In this work, we refer to paired scans that are not well aligned after registration (due to natural deformations of the abdomen) as imperfectly registered. In contrast to multimodal brain datasets, like BraTS[5], abdominal imaging faces the challenge of flexible and highly deformable anatomy[18]. Scans of the abdomen are often acquired by having subjects hold their breath to limit motion artifacts[19]. Some approaches for scanning the abdomen use a single breath hold, while other techniques exist that use two breath holds, creating two images by skipping slices, and interleaving them to obtain a single image volume covering the whole abdomen. For example, in this manuscript, we have T2w abdominal images taken on two interleaved breath holds and paired T1w images taken on a single breath hold. Because of the deformable characteristics of abdominal imaging, achieving perfectly aligned multimodal data is difficult, even after effective deformable registration (Fig 1).

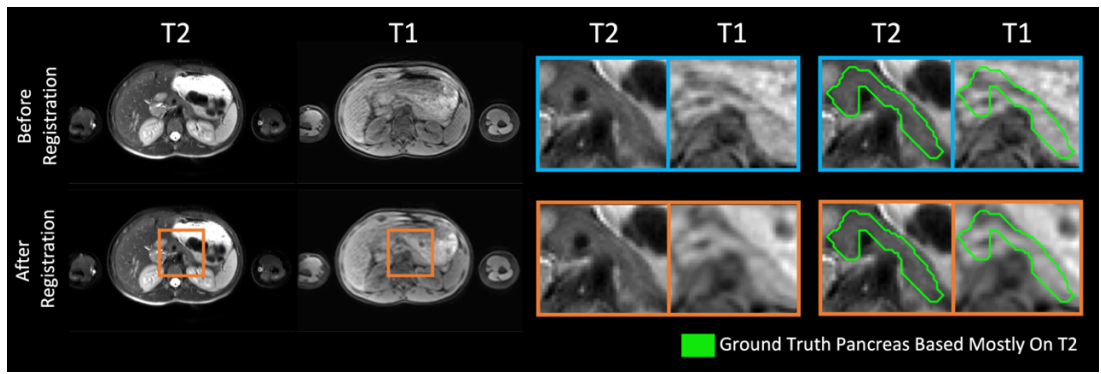

**Fig. 1** Our pancreas labels were mainly drawn based on the T2w MR images. Despite the use of effective deformable registration, the pancreas labels did not always line up well with the corresponding anatomy in the paired T1w MR images from the same subject and same imaging session. For example, the T1w image shown exhibits poor alignment with both the T2w image and the pancreas mask, even after registration (bottom right). Persisting image misalignment makes it unclear how best to fuse T1w information to see gains in segmentation performance. The example shown exhibits particularly poor alignment.

Although terminology can vary in the literature, fusion of medical images can be generally separated into early fusion, late fusion, and hybrid fusion[20]. These names are intuitive: early fusion means that a model combines multimodal information towards the beginning of processing, late fusion means information combination happens towards the end of processing, and hybrid fusion uses both early and late fusion.

Where in the deep learning model to perform multimodal medical image fusion is still an open question. Moreover, in the deformable anatomy of the abdomen, the best fusion approach for pancreas segmentation on multimodal MR images is unclear, especially because it is challenging to perfectly align anatomical structures across modalities. In response, we study 1) how the stage of fusion, from early through late, affects the quality of pancreas segmentation on imperfectly registered paired T2w and T1w MR images when using UNet[21]-based architectures and 2) whether there is a generalized "best" location to fuse in our problem space.

## 2 Methods

We examined fusion at all main intermediate stages in UNet-based networks (Figure 2). We first deformably registered the paired T2w and T1w abdominal MR images using deeds registration[22,23]. Then, we trained a set of basic UNets[21] with different fusion points to assess how early through late fusion influences pancreas segmentation on imperfectly registered data. We assessed whether our findings generalized to another model by validating on nnUNet[24].

*2.1 Data*

Samples were studied in deidentified form from Vanderbilt University Medical Center under Institutional Review Board approval (IRB #130883) and are associated with ClinicalTrials.gov identifier: NCT03585153. Our dataset contained 353 images pairs of T2w and T1w abdominal MR images which came from 163 subjects. Here, an image pair means a T1w and T2w image of the same subject acquired during the same imaging session. Scans with major motion artifacts were excluded. Each T2w image volume had a spatial resolution $1.5 \times 1.5 \times 5$ mm with spatial dimension of $256 \times 256 \times 30$ voxels. The T1w images were at a higher spatial resolution than the T2w images at $[1.18, 1.30] \times [1.18, 1.30] \times 2$ mm with spatial dimension of $[240, 320] \times 240 \times [75, 98]$ voxels. Each image pair had a corresponding manual delineation of the pancreas, which was drawn on the T2w image and based primarily on the T2w image information. Of the 163 subjects, 63 were controls, 66 had type 1 diabetes, 22 were autoantibody positive, and 12 had maturity-onset diabetes of the young (MODY). Most of this cohort were part of a previous study[10].

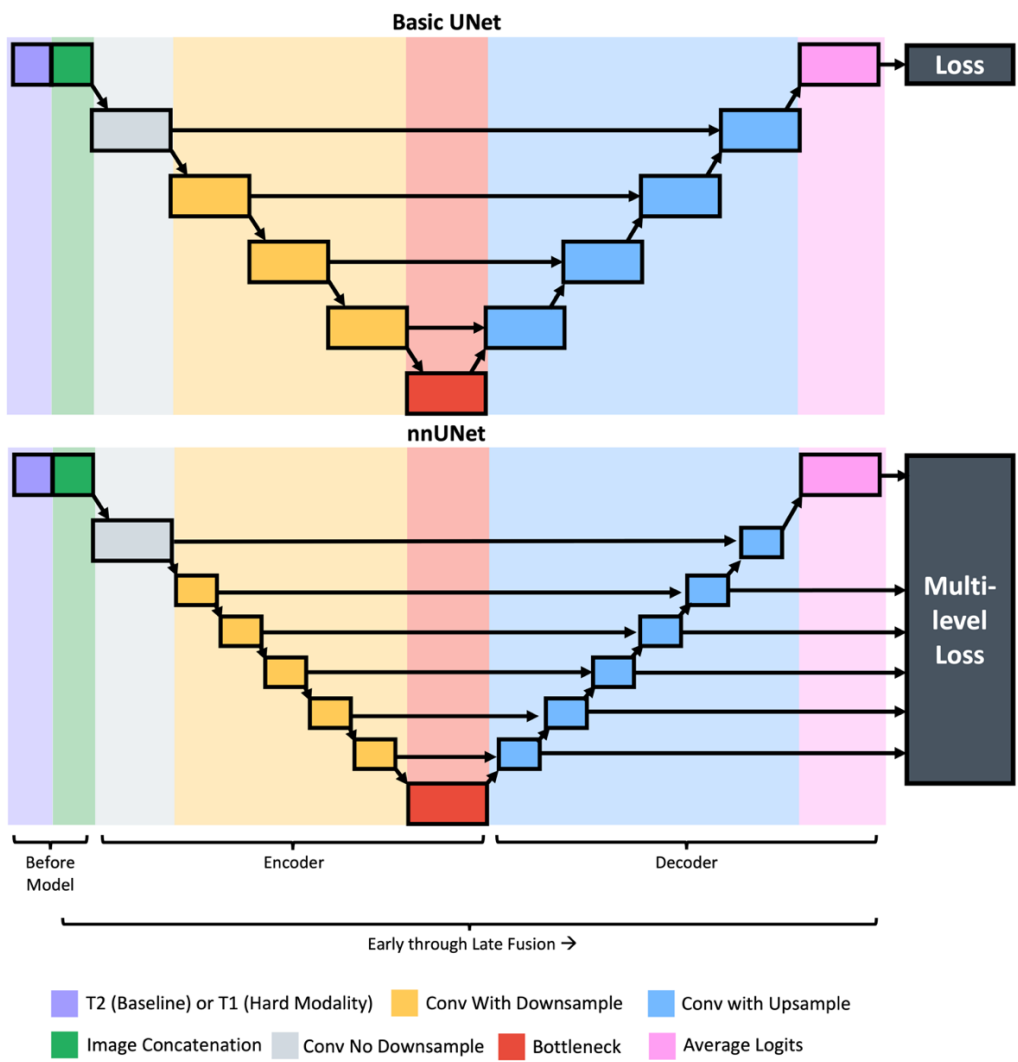

**Fig. 2** We studied how early through late fusion affects pancreas segmentation on imperfectly registered image pairs using UNet-based architectures. We used the MONAI basic UNet (top) and nnUNet (bottom) to investigate if there exists a generalizable "best" location to fuse for our problem domain. In this pursuit, we trained 13 basic UNet configurations and 17 nnUNet configurations, each with a single fusion point. All of the fusion points are shown above as colored blocks. All fusion was performed through concatenation, except for the latest fusion (pink) which used averaging logits.

*2.2 Registration*

The T2w/T1w image pairs were acquired during the same session, so the subject position and anatomy were not different between scans. However, each T2w image was an interleaving of two separate scans on two different breath holds, and each T1w image was taken on a separate single breath hold. This resulted in misalignment between the paired images, which necessitated deformable registration.

The T1w images were deformably registered to their paired T2w images using deeds registration[22,23]. Deeds is a state-of-the-art deformable image registration technique that was specifically designed to handle abdominal images. Because the output of deeds is resampled to isotropic resolution by default, we resampled the isotropic T1w image back into the original T2w image space. Each registered image pair was qualitatively assessed for alignment between T2w image, T1w image, and the pancreas mask. The registered images were generally well-aligned; however, the registration was not perfect (figure 1).

*2.3 UNet Fusion Configurations*

To assess how fusion location affects pancreas segmentation on the UNet-style architecture, we opted for a straightforward design. Each architecture variant had one fusion point—the blocks on which fusion occurred are shown in figure 2. At each major block, we had an architecture variant that fused through concatenation. Additionally, we trained a multimodal model that accepted T2w and T1w images that were fused in image space before being fed to the model. To handle very late fusion, we built an architecture that fused information by averaging logits. We also used single modality models as baselines: one for only T2w images, and another for only T1w images. These choices resulted in 13 model variants for the basic UNet from MONAI[25], and 17 variants for nnUNet.

*2.4 Cross-validation*

For 5-fold cross-validation, each split was performed at the subject level, and always stratified across conditions: control, type 1 diabetes, autoantibody positive, and MODY, to reduce bias. Each cross-validation fold was partitioned into three sets: training, validation, and test. The training and validation sets consisted of 90% and 10% of the non-test subjects for the fold, respectively. The selection of subjects per fold was random. The validation and testing data contained a single scan from each subject, while we allowed multiple scans per subject in the training data.

For the basic UNets, we performed cross-validation on each of the 13 fusion configurations. We trained each basic UNet architecture using the same approach. Each MR image that was loaded had the top 0.001% of pixel intensities clipped and was min-max normalized between zero and one based on its minimum and maximum voxel intensity. The training loop consisted of 16000 steps (batches) using a batch size of two image volumes, Dice loss[26], the Adam optimizer[27], the one cycle learning rate scheduler[28] with a max learning rate of 0.001, and early stopping. The model learned after each batch and performed validation every 20 steps. Early stopping engaged if the average validation Dice loss did not improve after 500 batches. The weights were selected based on the validation step with the lowest validation loss.

For the nnUNets, we used the default 5-fold cross-validation and weight selection approach from nnUNet version 2. We stipulated that the data in each fold and train/validation/test split matched with our basic UNet experiments.

*2.7 Technical Implementation*

All basic UNet experiments were implemented using Pytorch[29] 1.12.1 and were based on the Basic UNet architecture from MONAI[25] 1.0.0. All of the nnUNet experiments were performed and adapted from the nnUNet version 2 github: https://github.com/MIC-DKFZ/nnUNet and used their provided environment. The models were trained on NVIDIA RTX A6000s and NVIDIA Quadro RTX 5000s.

## 3 Results

*2.1 Performance Relative to Baseline*

As depicted in figure 3, across all the trained approaches (the T2w-only baseline, the imperfectly registered T1w-only model, and the fusion configurations), performance differences were relatively small in terms of the Dice score[30]. The T1w images were notably more challenging than the T2w images, which was not surprising, because the labels were drawn mainly on the T2w images, and the T1w images were imperfectly registered. Fusion of the T2w and T1w modalities led to both improvements and reductions in performance compared to baseline. Gains seen on key fusion points on the basic UNets did not transfer to nnUNet and vice versa. Improved performance from fusion on the basic UNet occurred in the middle of the encoder, at the bottleneck, in the middle of the decoder, and from averaging logits (late fusion). The nnUNet showed improved performance from very early fusion (image concatenation), with a steady decline in performance as fusion occurred later in the modeling process.

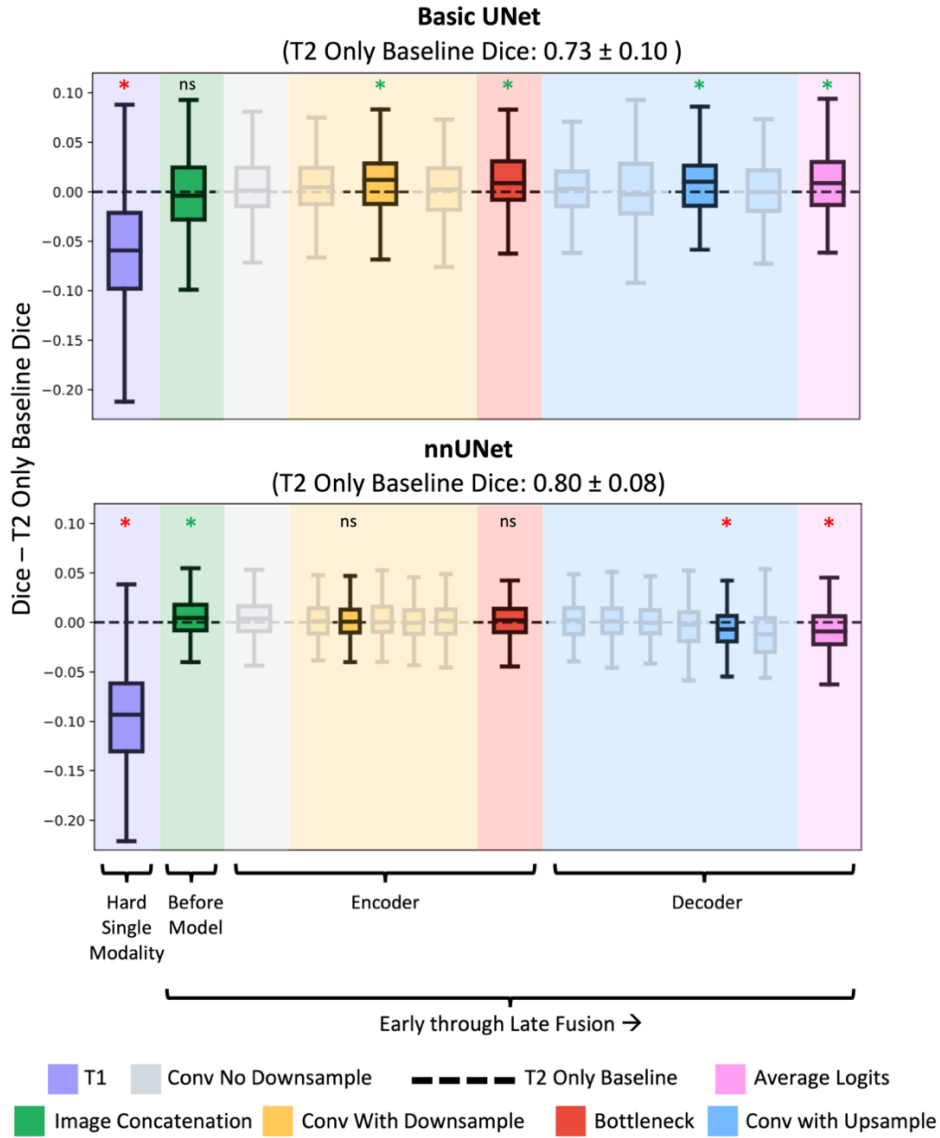

**Fig. 3** Fusion in specific blocks shows performance gains on both the basic UNet (top) and nnUNet (bottom), when compared to the T2w-only baseline (dotted line). However, these improved fusion points differ between the models and the gains are small. Key fusion points from the basic UNet are highlighted and transferred to nnUNet, as well as expected comparisons: T1w-only and naïve image concatenation. In the downsampling and upsampling portions of the models, correspondence between fusion points in the basic UNet and nnUNet was determined by counting $i$ layers from either the beginning or end of the section. Asterisks denote statistically significant results from the Wilcoxon signed rank test after Bonferroni correction, with green indicating improved performance and red indicating decreased performance. The models that are the focus of this figure are highlighted. All other fusion points are shown faded to summarize variability.

*2.2 Sensitivity Analysis*

In figure 4, we show that, when compared to the T2w-only baseline, the best fusion approaches generally improve performance on the diabetic pancreases that are smaller in size. Performance gains occur despite the difficulty of identifying these same smaller pancreases when using the T1w images alone. In figure 5, we show that the systemic bias for fusion was small—the nnUNet models depicted more agreement with the baseline than the basic UNets.

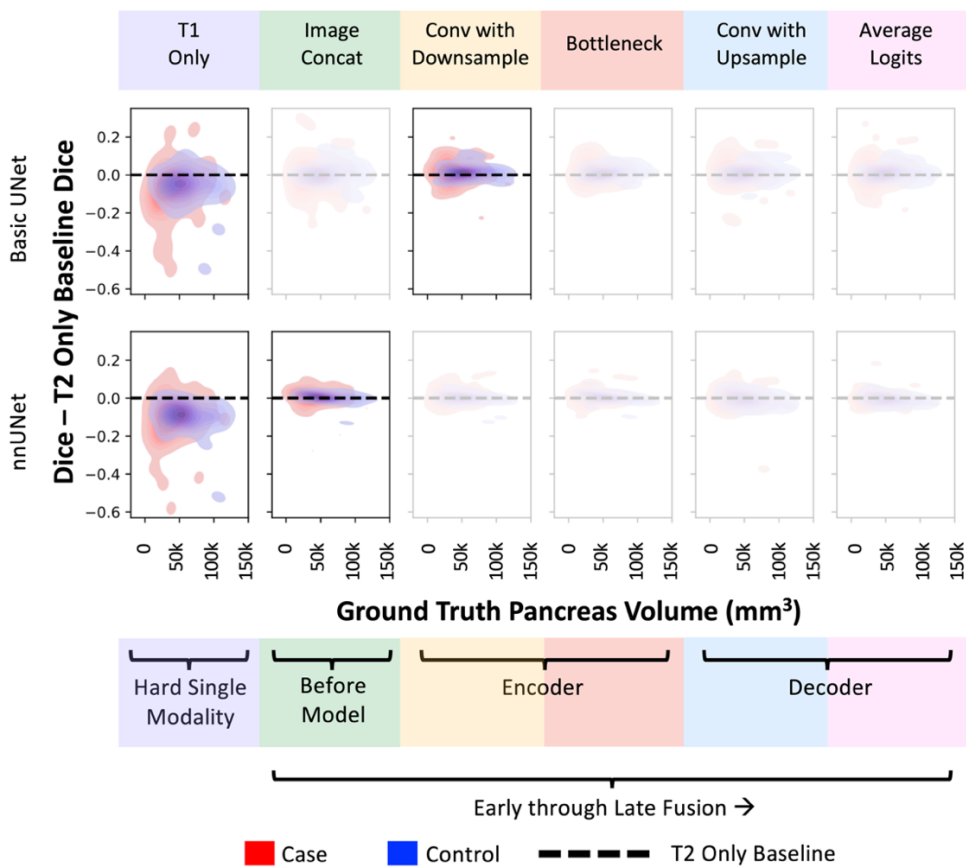

**Fig. 4** Generally, when our key fusion models performed better than the T2w-only baseline, it was more pronounced on smaller pancreases, which were usually from diabetic subjects (denoted as red contours). We highlight the T1w-only baseline in addition to the best fusion approach for both the basic UNet and nnUNet to demonstrate how fusing the more difficult T1w modality can improve performance compared to the T2w-only baseline (dotted line). Additional key fusion points are shown faded to summarize variability across fusion points.

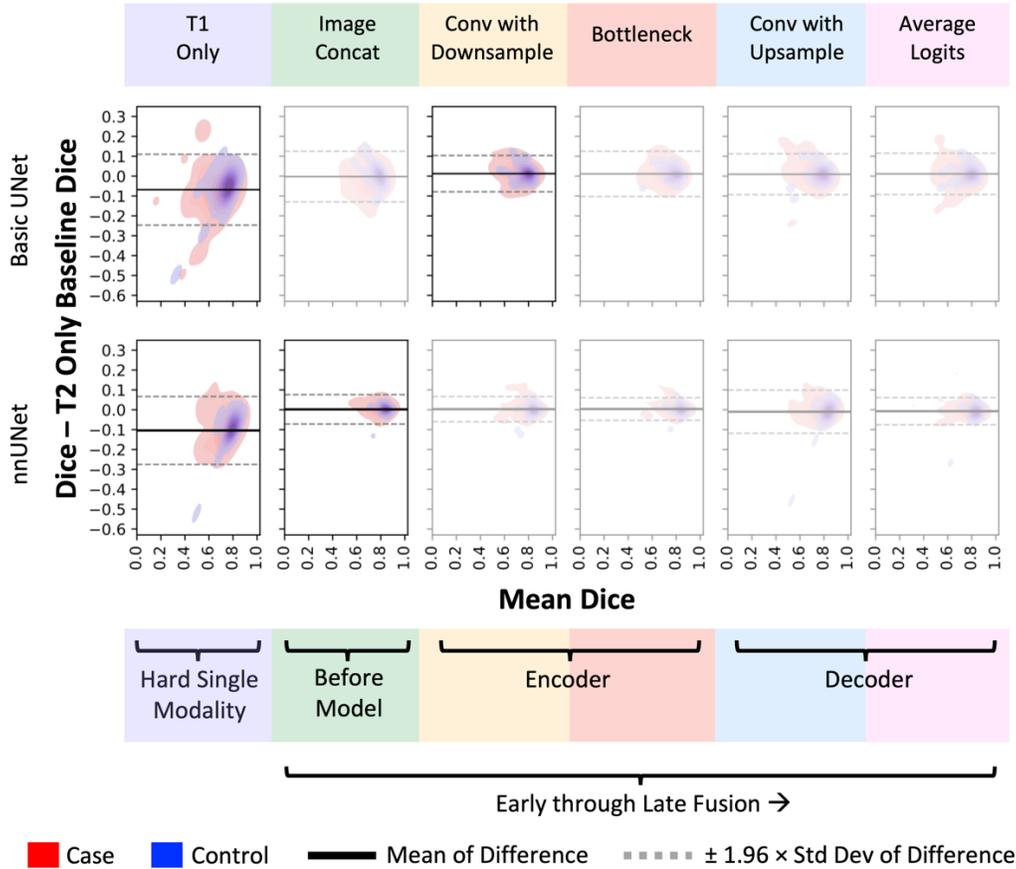

**Fig. 5** In this Bland Altman plot[31], we note that for the key fusion points 1) systemic biases were small and 2) the nnUNet configurations had greater agreement with the T2w-only baseline. The models that are the focus of this figure are highlighted (T1w only and best configuration for each model). Additional key fusion points are shown faded to summarize variability.

*2.3 Qualitative Results*

In figure 6, we visualized how the best fusion approach from the basic UNets compared to the T2w-only baseline and T1w-only model. By comparing the fusion to the single modality models, we noted three intuitive scenarios. When the T1w image provided signal that assisted in pancreas identification, performance could improve. When the T1w image did not provide helpful signal, the performance could be similar to the T2w-only baseline. When the T1w image presented information that made segmentation highly error prone, fusion could reduce performance compared to the baseline.

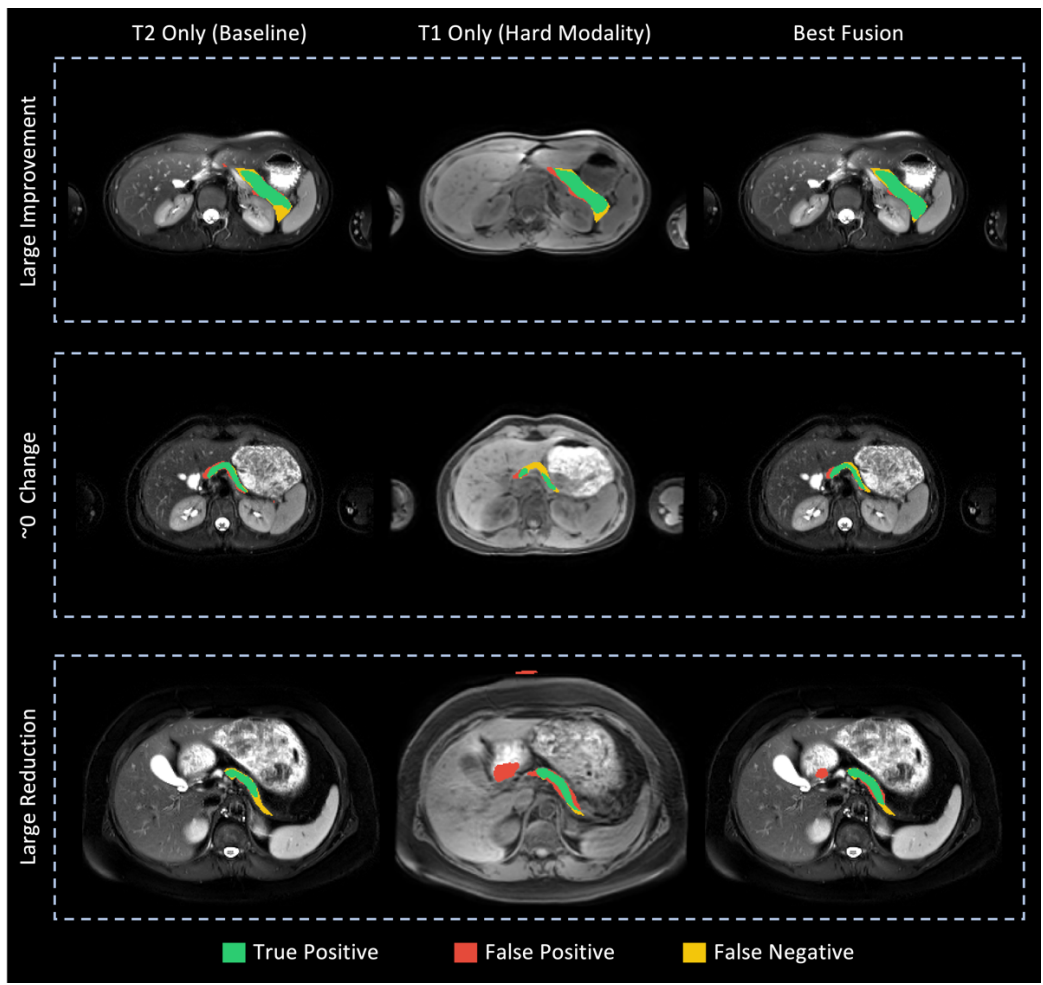

**Fig. 6** When the imperfectly registered T1w image was complementary to the T2w image, we saw improved performance through our best fusion (top row). If the T1w image did not complement the T2w image, fusion performance could be similar to baseline (middle row). When the T1w image was challenging for segmentation, fusion could reduce performance (bottom row).

## 4 Discussion & Conclusion

In this work, we designed experiments to discover if there were optimal and generalizable fusion points on UNet-based segmentation architectures when dealing with imperfectly registered multimodal data. The main finding was that improvements on fusion were small and model specific, rather than generalizable, even between the similar basic UNet and nnUNet architectures. These results indicate that further innovation may be needed to better capitalize on the complementary information in imperfectly registered image pairs.

**Disclosures**

The authors do not have disclosures or conflicts of interest.

**Code and Data Availability**

Access to code from this manuscript can be found at the following link: https://github.com/MASILab/influence_of_fusion_on_pancreas_segmentation. The data that support the findings of this article are not publicly available.

*Acknowledgments*

This work was supported by Integrated Training in Engineering and Diabetes, grant number T32 DK101003. This material is partially supported by the National Science Foundation Graduate Research Fellowship under Grant No. DGE-1746891. Research support was received from the Leona M. and Harry B. Helmsley Charitable Trust (2207-05374), JDRF International (3-SRA-2015-102-M-B and 3-SRA-2019-759-M-B), and the National Institute of Diabetes and Digestive and Kidney Diseases (DK 129979). This work used REDCap and VCTRS resources, which are supported by grant UL1 TR000445 from National Center for Advancing Translational Sciences, National Institutes of Health (NIH). Support was also received from the Vanderbilt Diabetes Research and Training Center, Division of Diabetes, Endocrinology, and Metabolic Diseases (DK 020593), and the Vanderbilt University Institute of Imaging Science Center for Human Imaging (1 S10OD021771 01). This work was supported by NIH R01DK135597 (Huo). This work was supported by NSF career 1452485 and NSF 2040462, and National Science Foundation (NSF) grant 2220401. This work was conducted in part using the resources of the Advanced Computing Center for Research and Education at Vanderbilt University, Nashville, TN. The Vanderbilt Institute for Clinical and Translational Research (VICTR) is funded by the National Center for Advancing Translational Sciences (NCATS) Clinical Translational Science Award (CTSA) Program, Award Number 5UL1TR002243-03. The content is solely the responsibility of the authors and does not necessarily represent the official views of the NIH. This work was supported by U54DK134302 (NIH) and U54EY032442 (NIH). We extend gratitude to NVIDIA for their support by means of the NVIDIA hardware grant. We have used AI as a tool in the creation of this content, however, the foundational ideas, underlying concepts, and original gist stem directly from the personal insights, creativity, and intellectual effort of the author(s). The use of generative AI serves to enhance and support the author's original contributions by assisting in the ideation, drafting, and refinement processes. All AI-assisted content has been carefully reviewed, edited, and approved by the author(s) to ensure it aligns with the intended message, values, and creativity of the work.

*References*

[1] Kalamkar, S. and A., G. M., "Multimodal image fusion: A systematic review," Decision Analytics Journal **9**, 100327 (2023).
[2] Hermessi, H., Mourali, O. and Zagrouba, E., "Multimodal medical image fusion review: Theoretical background and recent advances," Signal Processing **183**, 108036 (2021).
[3] Mohsen, F., Ali, H., El Hajj, N. and Shah, Z., "Artificial intelligence-based methods for fusion of electronic health records and imaging data," Sci Rep **12**(1), 17981 (2022).
[4] Polinati, S., Bavirisetti, D. P., Rajesh, K. N. V. P. S., Naik, G. R. and Dhuli, R., "The Fusion of MRI and CT Medical Images Using Variational Mode Decomposition," Applied Sciences **11**(22), 10975 (2021).
[5] Menze, B. H., Jakab, A., Bauer, S., Kalpathy-Cramer, J., Farahani, K., Kirby, J., Burren, Y., Porz, N., Slotboom, J., Wiest, R., Lanczi, L., Gerstner, E., Weber, M.-A., Arbel, T., Avants, B. B., Ayache, N., Buendia, P., Collins, D. L., Cordier, N., et al., "The Multimodal Brain Tumor Image Segmentation Benchmark (BRATS)," IEEE Trans Med Imaging **34**(10), 1993–2024 (2015).
[6] Cole, J. B. and Florez, J. C., "Genetics of diabetes mellitus and diabetes complications," Nat Rev Nephrol **16**(7), 377–390 (2020).
[7] Gurney, J. K., Stanley, J., York, S., Rosenbaum, D. and Sarfati, D., "Risk of lower limb amputation in a national prevalent cohort of patients with diabetes," Diabetologia **61**(3), 626–635 (2018).
[8] "IDF Diabetes Atlas 10th edition.", (2021).
[9] Marshall, S. M., "The pancreas in health and in diabetes," Diabetologia **63**(10), 1962–1965 (2020).
[10] Virostko, J., Williams, J., Hilmes, M., Bowman, C., Wright, J. J., Du, L., Kang, H., Russell, W. E., Powers, A. C. and Moore, D. J., "Pancreas Volume Declines During the First Year After Diagnosis of Type 1 Diabetes and Exhibits Altered Diffusion at Disease Onset," Diabetes Care **42**(2), 248–257 (2019).
[11] Goda, K., Sasaki, E., Nagata, K., Fukai, M., Ohsawa, N. and Hahafusa, T., "Pancreatic volume in type 1 und type 2 diabetes mellitus," Acta Diabetol **38**(3), 145–149 (2001).
[12] Macauley, M., Percival, K., Thelwall, P. E., Hollingsworth, K. G. and Taylor, R., "Altered Volume, Morphology and Composition of the Pancreas in Type 2 Diabetes," PLoS One **10**(5), e0126825 (2015).
[13] Philippe, M.-F., Benabadji, S., Barbot-Trystram, L., Vadrot, D., Boitard, C. and Larger, E., "Pancreatic Volume and Endocrine and Exocrine Functions in Patients With Diabetes," Pancreas **40**(3), 359–363 (2011).
[14] Yao, X., Song, Y. and Liu, Z., "Advances on pancreas segmentation: a review," Multimed Tools Appl **79**(9–10), 6799–6821 (2020).
[15] Ghorpade, H., Jagtap, J., Patil, S., Kotecha, K., Abraham, A., Horvat, N. and Chakraborty, J., "Automatic Segmentation of Pancreas and Pancreatic Tumor: A Review of a Decade of Research," IEEE Access **11**, 108727–108745 (2023).
[16] Cai, J., Lu, L., Zhang, Z., Xing, F., Yang, L. and Yin, Q., "Pancreas Segmentation in MRI Using Graph-Based Decision Fusion on Convolutional Neural Networks," 442–450 (2016).
[17] El-Gamal, F. E.-Z. A., Elmogy, M. and Atwan, A., "Current trends in medical image registration and fusion," Egyptian Informatics Journal **17**(1), 99–124 (2016).

[18] Xu, Z., Lee, C. P., Heinrich, M. P., Modat, M., Rueckert, D., Ourselin, S., Abramson, R. G. and Landman, B. A., "Evaluation of Six Registration Methods for the Human Abdomen on Clinically Acquired CT," IEEE Trans Biomed Eng **63**(8), 1563–1572 (2016).

[19] Gdaniec, N., Eggers, H., Börnert, P., Doneva, M. and Mertins, A., "Robust abdominal imaging with incomplete breath-holds," Magn Reson Med **71**(5), 1733–1742 (2014).

[20] Gadzicki, K., Khamsehashari, R. and Zetzsche, C., "Early vs Late Fusion in Multimodal Convolutional Neural Networks," 2020 IEEE 23rd International Conference on Information Fusion (FUSION), 1–6, IEEE (2020).

[21] Ronneberger, O., Fischer, P. and Brox, T., "U-Net: Convolutional Networks for Biomedical Image Segmentation," 234–241 (2015).

[22] Heinrich, H. P., Jenkinson, M., Brady, M. and Schnabel, J. A., "MRF-Based Deformable Registration and Ventilation Estimation of Lung CT," IEEE Trans Med Imaging **32**(7), 1239–1248 (2013).

[23] Heinrich, M. P., Maier, O. and Handels, H., "Multi-modal Multi-Atlas Segmentation using Discrete Optimisation and Self-Similarities."

[24] Isensee, F., Jaeger, P. F., Kohl, S. A. A., Petersen, J. and Maier-Hein, K. H., "nnU-Net: a self-configuring method for deep learning-based biomedical image segmentation," Nat Methods **18**(2), 203–211 (2021).

[25] Cardoso, M. J., Li, W., Brown, R., Ma, N., Kerfoot, E., Wang, Y., Murrey, B., Myronenko, A., Zhao, C., Yang, D., Nath, V., He, Y., Xu, Z., Hatamizadeh, A., Myronenko, A., Zhu, W., Liu, Y., Zheng, M., Tang, Y., et al., "MONAI: An open-source framework for deep learning in healthcare" (2022).

[26] Sudre, C. H., Li, W., Vercauteren, T., Ourselin, S. and Jorge Cardoso, M., "Generalised Dice Overlap as a Deep Learning Loss Function for Highly Unbalanced Segmentations," 240–248 (2017).

[27] Kingma, D. P. and Ba, J., "Adam: A Method for Stochastic Optimization" (2014).

[28] Smith, L. N. and Topin, N., "Super-Convergence: Very Fast Training of Neural Networks Using Large Learning Rates" (2017).

[29] Paszke, A., Gross, S., Massa, F., Lerer, A., Bradbury, J., Chanan, G., Killeen, T., Lin, Z., Gimelshein, N., Antiga, L., Desmaison, A., Köpf, A., Yang, E., DeVito, Z., Raison, M., Tejani, A., Chilamkurthy, S., Steiner, B., Fang, L., et al., "PyTorch: An Imperative Style, High-Performance Deep Learning Library" (2019).

[30] Dice, L. R., "Measures of the Amount of Ecologic Association Between Species," Ecology **26**(3), 297–302 (1945).

[31] Martin Bland, J. and Altman, DouglasG., "STATISTICAL METHODS FOR ASSESSING AGREEMENT BETWEEN TWO METHODS OF CLINICAL MEASUREMENT," The Lancet **327**(8476), 307–310 (1986).

**Biography**

Lucas W. Remedios is a PhD student in Computer Science at Vanderbilt University. His research focuses on artificial intelligence in medical imaging domains. He is advised by Dr. Bennett Landman.

Other author biographies are not available.